%% file: ms.tex
\documentclass[10pt,twocolumn,letterpaper]{article}

\usepackage{cvpr}
\usepackage{times}
\usepackage{epsfig}
\usepackage{graphicx}
\usepackage{amsmath}
\usepackage{amssymb}
\usepackage{algorithm}
\usepackage{adjustbox}
\usepackage{import}
\usepackage[noend]{algpseudocode}

\usepackage[pagebackref=true,breaklinks=true,letterpaper=true,colorlinks,bookmarks=false]{hyperref}

\cvprfinalcopy 

\def\cvprPaperID{5721} 

\begin{document}

\title{DeepCaps: Going Deeper with Capsule Networks}

\author{Jathushan Rajasegaran$^{1}$\\
\and
Vinoj Jayasundara$^{1}$ \\
\and
Sandaru Jayasekara$^{1}$ \\
\and
Hirunima Jayasekara$^{1}$ \\
\and
Suranga Seneviratne$^{2}$
\and
Ranga Rodrigo$^{1}$ \\
\and
$^1$Department of Electronic and Telecommunication Engineering, University of Moratuwa\\
$^2$School of Computer Science, University of Sydney\\
\tt\small{\{brjathu, vinojjayasundara, sandaruamashan, nhirunima\}@gmail.com} \\
\tt\small{suranga.seneviratne@sydney.edu.au, ranga@uom.lk}
}
\maketitle
\begin{abstract}
\import{/}{abstract}
\end{abstract}
\section{Introduction}
\import{/}{introduction}

\section{Related Work}

\import{/}{Related_work}
\label{related_works}

\section{DeepCaps}
\label{Deep_Caps}
\import{/}{methods}

\subsection{3D Convolution Based Dynamic Routing}
\import{/}{methods_3d_conv}

\subsection{DeepCaps Architecture}
\import{/}{methods_deep_archi}

\subsection{Loss Function}
\import{/}{methods_deep_loss}


\section{Class Independent Decoder Network}
\label{decoder_net}

\import{/}{methods_decoder}

\section{Experiments and Results}
\label{reslts}

\subsection{Implementation}
\import{/}{implementation}

\import{/}{results}

\import{/}{conclusion}

\label{conclusion}

\section{Acknowledgement}
The authors thank National Research Council, Sri Lanka  (Grant 12-018), and the Faculty of Information Technology of the University of Moratuwa, Sri Lanka for providing computational resources.
{\small
\bibliographystyle{IEEEtranS}
\bibliography{egbib}
}

\end{document}

%% file: abstract.tex
Capsule Network is a promising concept in deep learning, yet its true potential is not fully realized thus far, providing sub-par performance on several key benchmark datasets with complex data. Drawing intuition from the success achieved by Convolutional Neural Networks (CNNs) by going deeper, we introduce DeepCaps\footnote{\href{https://github.com/brjathu/deepcaps}{https://github.com/brjathu/deepcaps}}, a deep capsule network architecture which uses a novel 3D convolution based dynamic routing algorithm. With DeepCaps, we surpass the state-of-the-art results in the capsule network domain  on CIFAR10, SVHN and Fashion MNIST, while achieving a 68\% reduction in the number of parameters. Further, we propose a class-independent decoder network, which strengthens the use of reconstruction loss as a regularization term. This leads to an interesting property of the decoder, which allows us to identify and control the physical attributes of the images represented by the instantiation parameters.

\vspace*{-0.6cm}

%% file: introduction.tex

\vspace*{-0.1cm}
In the last few years, convolutional neural networks (CNNs) made breakthroughs in many computer vision tasks, and significantly outperformed many conventional curated feature driven models. Two common themes of increasing the performance of CNNs are to increase the depth and the width of the network (\eg, the number of levels of the network and the number of units at each level) and to use as much training data as possible. Although CNNs have been successful, they have few limitations such as the invariance caused by pooling and the inability to understand spatial relationship between features. To address these limitations, Sabour \etal proposed Capsule Networks~\cite{sabour2017dynamic} which have shown promising comparable results to CNNs in several standard datasets. Intuitively, attempting to go deeper with capsule networks is a step in the right direction to further enhance their performance.


The capsule network (CapsNet) model proposed by Sabour \etal ~\cite{sabour2017dynamic} comprises only one convolution layer and one fully-connected capsule layer. The proposed architecture works well with the MNIST~\cite{mnist} dataset, nonetheless the performance on datasets with more complex objects such as CIFAR10~\cite{cifar10} is not on par with the CNNs, due to the nature of complex shapes in CIFAR10 in comparison to MNIST.

A naive attempt of creating a deep CapsNet by simply stacking such fully-connected capsule layers will result in an architecture similar to a MLP model which has several limitations. First, dynamic routing used in capsule networks is an extremely computationally expensive procedure, and having multiple routing layers incur higher costs of training and inference time. Second, it has been recently shown that stacking fully connected capsule layers on top of each other will result in poor learning in the middle layers~\cite{xi2017capsule}. This is due to the fact that when there are too many capsules, the coupling coefficients tend to be too small, consequently dampening the gradient flow and inhibiting learning. Third, it has been shown that, especially in the lower layers, correlated units tend to concentrate in local regions \cite{szegedyc}. Although localized routing can conspicuously take advantage of this observation, such localized routing cannot be implemented in fully connected capsules.

In order to address these limitations caused by stacking capsule layers, we propose the following solutions. To reduce the computational complexity introduced by multiple layers needing dynamic routing, several avenues are possible: Reducing the number of routing iterations in the initial layers that are larger in size reduces the complexity while not affecting the features as they need not be complex in nature. In addition, using 3D-convolution-inspired routing in the middle layers --due to parameter sharing-- reduces the number of parameters. We can address the problem of poor learning in the middle layers due to naive stacking by improving the gradient flow, that involve skip connections coupled with convolutions. Moreover, while reducing the complexity, the deep capsule network must be able to handle richer data sets than MNIST. We propose that localized routing will be able to capture the higher level information better than fully connected routing.

Sabour \etal ~\cite{sabour2017dynamic} used regularization through the incorporation of reconstruction error (which is generated by the decoder network) to reduce over fitting. Nevertheless, a stronger regularization than \cite{sabour2017dynamic} is necessary to reduce overfitting when developing deeper networks, due to the inherent increase in the model complexity with model depth. Hence, in an attempt to enhance the regularization, we propose a class-independent decoder. We observed an interesting property of this decoder, which provides controllability over the learning and perturbation of instantiation parameters. In existing capsule networks and decoders, it is not possible to guarantee that the physical property represented by a given instantiation parameter is the same across all the classes. In the proposed decoder, it is guaranteed that the represented property will be the same for any given instantiation parameter across all the classes, providing higher controllability, which is immensely advantageous in practical applications and theoretical studies.


To this end, in this paper, we propose {\bf DeepCaps}: a deep capsule network architecture by leveraging two key ideas: \emph{Dynamic routing} and \emph{Going deeper in the network}. The novel dynamic routing algorithm that we propose achieves parameter reduction and localized routing, making the routing possible in a convolutional framework rather than resorting to fully-connected capsules, while skip connections allow us to train deeper networks. More specifically, we make the following contributions in the paper:

\begin{itemize}

\item Proposing a novel deep architecture for capsule networks, termed {\bf DeepCaps}, that aims at improving the performance of the capsule networks for more complex image datasets. Further, we propose a novel 3D-convolution-based dynamic routing algorithm to aid the learning process of DeepCaps.


\item Proposing a novel class-independent decoder network, which acts as a better regularization term. We further investigate on the observation that this novel decoder has the ability to provide controllability over the instantiation parameters. 

\item Evaluating the performance of {\bf DeepCaps} on several benchmark datasets: We significantly outperform the existing state-of-the-art capsule network architectures, while requiring a significantly lower number of parameters. For example, for the CIFAR10 dataset, DeepCaps achieves a 3\% improvement in the accuracy in comparison to \cite{sabour2017dynamic}, along with a 68\% reduction in the number of parameters.


\end{itemize}

The rest of the paper is organized as follows: In Section \ref{related_works}, we discuss the related work on Capsule Networks, Section \ref{Deep_Caps} describes our DeepCaps architecture and the novel 3D routing algorithm, Section \ref{decoder_net} outlines the class-independent decoder network. Section \ref{reslts} shows our results. Finally, Section \ref{conclusion} concludes the paper.




%% file: Related_work.tex
One of the major issues which we face with deep networks is the vanishing/exploding gradients. When the error signal 
is passed through many layers, it can vanish and “wash out” by the time it reaches the beginning of the network \cite{cortes2016adanet}, \cite{hariharan2015hypercolumns}, which hinders the convergence. This issue is being addressed in many models proposed, where ResNets \cite{he2016deep} and Highway Networks \cite{srivastava2015training}  bypass signals from one layer to the next via identity connections. Stochastic depth \cite{huang2016deep} shortens ResNets by randomly dropping layers during training to allow better information and gradient flow.
DenseNets \cite{huang2017densely} ensure the maximum information flow between layers in the network, by connecting all layers (with matching feature-map sizes) directly with each other. To preserve the feed-forward nature, each layer obtains additional inputs from all preceding layers and passes on its own feature-maps to all subsequent layers. They create short paths from early layers to latter layers.

The idea of grouping the neurons is proposed in Hinton \etal \cite{hinton2011transforming}. As an extension to this, Sabour \etal ~\cite{sabour2017dynamic} proposed a dynamic routing algorithm between capsules, using the concept of routing by agreement between capsules. Dynamic routing helps the network to achieve eqivarience, where CNNs can only achieve in-variance by the pooling operation. In addition to dynamic routing, Hinton \etal \cite{hinton2018matrix} used EM routing for matrix capsules representing each entity by a pose matrix. There have been many extensions to these: HitNet \cite{deliege2018hitnet} uses a hybrid hit and miss layer for data augmentations. Dilin \etal \cite{wang2018optimization} solves the dynamic routing as an optimization problem, and achieves better performance by introducing KL divergence between the coupling distribution. CapsGan \cite{jaiswal2018capsulegan} uses a capsule network as the discriminator in the GAN pipeline, to get visually better results than convolutional GANs.  In contrast to these, our work focuses on going deeper with the capsule networks and increase its performance on more complex datasets. 

SegCaps \cite{lalonde2018capsules} uses capsules for image segmentation and they achieve the state-of-the-art results on LUNA16 dataset. This is the closest work to ours on the basis of routing. They use 2D convolution for the voting procedure. By using 2D convolutions, it takes all the capsule along depth as the inputs for the transformation, thus, mixing the information contained in the capsules. In our 3D-convolution-based routing, we design the strides along the depth to be the capsule dimension, as a result of which, each capsules along the depth dimension is voted separately.

Our work explores the possibilities of creating deeper networks consisting of multiple capsule layers. We believe, to the best of our knowledge, that this is the first attempt to go deeper with capsule networks. Further, the instantiation parameters of the capsule networks have shown a novel way of representing the images, by encoding physical variations such as rotation and skewness in a vector. A small perturbation in an instantiation parameter will change the corresponding physical variations in the reconstructed image. Still which parameter causes what kind of changes in the reconstructed images has been not studied.

\vspace*{-0.15cm}

%% file: methods.tex


One of the main drawbacks with dynamic routing in the current form~\cite{sabour2017dynamic} is that it can only be implemented in a fully connected manner (\eg, it cannot be implemented in a convolutional manner).  In~\cite{sabour2017dynamic}, after the primary capsule layer, capsule vectors are flattened and dynamically routed to the classification capsules. Thus, if it is necessary to go deep into the architecture with the dynamic routing algorithm in \cite{sabour2017dynamic}, we need to keep stacking fully connected capsule layers, which is equivalent to stacking fully connected layers in MLP models. This is not computationally efficient as the feature space is large at the start of the network. Hence, in order to stack convolutional capsule layers similar to the conventional CNNs, a novel dynamic routing algorithm is required.



%% file: methods_3d_conv.tex
Let the output of the capsule layer $l$ be $\mathbf{\Phi}^{l} \in \mathbb{R}^{(w^{l},w^{l},c^{l},n^{l})}$, where $w^{l}$ is the  height and the width of the feature map, $c^{l}$ is the number of 3D capsule tensors, and $n^{l}$ is the number of atoms (i.e. capsule dimension). In this section, we illustrate the novel mechanism that we propose in order to route the 3D capsule tensors from layer $l$ to predict the new 3D capsule tensor $\mathbf{\Phi}^{l+1} \in \mathbb{R}^{(w^{l+1},w^{l+1},c^{l+1},n^{l+1})}$. 

\begin{figure*}[!h]
    \centering
    \includegraphics[scale=0.1, angle=-90]{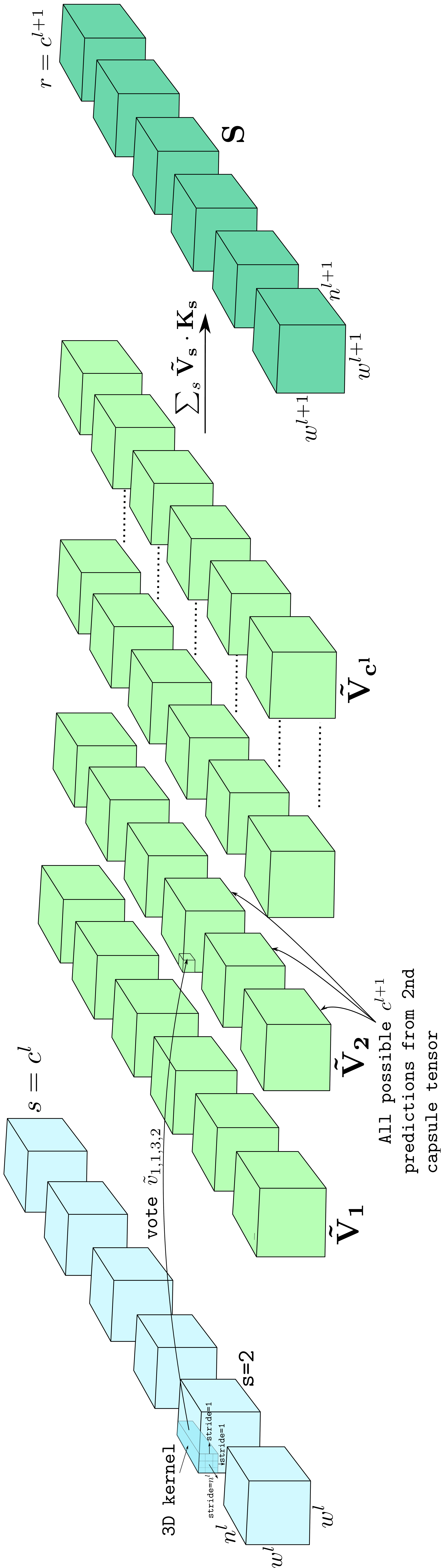}
    \vspace{0.2cm}
    \caption{Dynamic routing using 3D convolutions: In a high level explanation, each capsule tensor in layer $l$ will predict $c^{l+1}$ capsule tensors. Therefore, $c^l$ number of predictions are available for a capsule tensor in layer $l+1$. In the first routing iteration, all are equally weighted and summed together to get the final prediction $\mathbf{S}$. Then, in the following iterations, coupling coefficients are updated according to the agreement with $\mathbf{S}$ and $\mathbf{\Tilde{V}}$.}
    \label{fig:3d}
\end{figure*}

First, we reshape $\mathbf{\Phi}^{l}$ into a single channel tensor $\mathbf{\Tilde{\Phi}}^{l}$, which has a shape of $(w^{l},w^{l},c^{l} \times n^{l},1)$ and convolve it with ($c^{l+1} \times n^{l+1}$) number of 3D convolutional kernels. Let $\mathbf{\Psi}^l_t$ be the $t^{th}$ kernel in layer $l$ where $t \in [c^{l+1} \times n^{l+1}]$, which results in the intermediate votes $\mathbf{V}$, and has the shape of $(w^{l+1},w^{l+1},c^{l}, c^{l+1} \times n^{l+1})$. Keeping the size of $\mathbf{\Psi}^l_t$ and stride as $n^l$ along with depth, allow us to get a vote for single capsule from layer $l$. See Fig. \ref{fig:3d}. Using a 3D convolution kernel with height and the width of the kernel greater than 1 as the transformation matrix, allows us to predict higher level capsules using a set of lower level capsules. 

Each element $v_{i,j,k,m}$ in $\mathbf{V}$ can be obtained by performing the 3D convolution operation, which is defined according to the Eq.~\ref{eq2} below:

\begin{equation}
\label{eq2}
v_{i,j,k,m} = \sum_p \sum_q \sum_r \mathbf{\Tilde{\Phi}}^{l}(i-p,j-q,k-r) \cdot \mathbf{\Psi}^l_t(p,q,r)  \ \ \ \
\end{equation}

In order to keep the shape of the intermediate votes $\mathbf{V}$ to be consistent with number of channels in the input capsule tensor $\mathbf{\Tilde{\Phi}}^{l}$, we use $(1,1,n^{l})$ as the stride for the 3D convolution operations.


Subsequently, we reshape the intermediate votes $\mathbf{V}$ to the inceptive votes  $\mathbf{\Tilde{V}}$ for the proposed iterative routing algorithm. It has the shape of $(w^{l+1},w^{l+1},n^{l+1}, c^{l+1} , c^{l})$, since we are predicting $c^{l+1}$ capsule tensors for each $s \in c^{l}$. Here, the value of $w^{l+1}$ can be analytically calculated using the Eq. \ref{eq1} below:

\begin{equation}
\label{eq1}
w^{l+1} = \frac{w^{l} - \texttt{Kernel size} + 2\times \texttt{Padding} }{\texttt{Stride}} + 1
\end{equation}

If the dynamic routing algorithm in~\cite{sabour2017dynamic} was used for routing, it would have routed all capsules in layer $l$ to all the capsules in layer $l+1$. However, the feature maps resulting from the convolution operation have localized features, thus, adjacent capsules share similar information. We can eliminate this redundancy by routing a block of capsules $s$, from layer $l$ to the capsules in layer $l+1$, instead of routing each capsule in layer $l$ individually. This modification results in a significant reduction of the number of parameters by a factor $c \cdot (w^l w^{l+1})^2$, in comparison to the dynamic routing algorithm. 

Similarly, with a 3D convolutional kernel transforming a subset of capsules in a block to one vote, we achieve localized voting. For example, a $3 \times 3 \times 8$ kernel will transform the adjacent 9 capsules to one vote. In other words, in layer $l$, a low level entity may be represented by either a single capsule, or more often a group of capsules, which are adjacent to each other. Hence, rather than routing them separately to a higher level capsule, we group them together and route. Due to these additional requirements that are not fulfilled by the existing routing algorithms, we propose the following novel routing algorithm.

First, we initialize the logits $\mathbf{B}_s$ as $\mathbf{0}$, where $\mathbf{B}_s \in \mathbb{R}^{(w^{l+1},w^{l+1},c^{l+1})}$, for each $s \in [c^{l}]$. The corresponding coupling coefficients $\mathbf{K}_{s}$ are calculated using a $\texttt{softmax\_3D}$ function, as defined by Eq. \ref{eq:eq4}, which we propose as a 3D version of the existing softmax function. \cite{sabour2017dynamic} 

\begin{equation}
\begin{aligned}\label{eq:eq4}
\mathbf{K}_s &= \texttt{softmax\_3D}(\mathbf{B}_s) \\
k_{pqrs} &= \frac{\exp(b_{pqrs})}{\sum_x \sum_y \sum_z \exp(b_{xyzs})}
\end{aligned}
\end{equation}

Here, the logits are normalized among all the predicted capsules from capsule tensor $s$ in layer $l$. This is due to the fact that a single capsule tensor in layer $l$ predicts all the possible outputs of every $(p,q,r)^{th}$ capsule in the layer $l+1$. In other words, each capsule tensor in layer $l+1$ will have $c^{l}$ corresponding predictions from layer $l$. Each prediction will be weighted with $k_{pqrs}$ to get a single prediction $S_{pqr}$, which will be passed through $\texttt{squash\_3D}$ function, as defined by Eq. \ref{eq:eq5}, to limit the length of a capsule vector between 0 and 1, as it represents the probability of existence of an entity.

\begin{equation}
\begin{aligned}\label{eq:eq5}
\hat{S}_{pqr} &= \texttt{squash\_3D}(S_{pqr}) \\
    &= \frac{\|S_{pqr}\|^2}{1 + \|S_{pqr}\|^2} \cdot \frac{S_{pqr}}{\|S_{pqr}\|}
\end{aligned}
\end{equation}

The key concept of the routing algorithm proposed by~\cite{sabour2017dynamic} is routing by agreement between the outputs of the capsules. The agreement between $\mathbf{\hat{S}}$ and $\mathbf{\Tilde{V}}_s$ is measured by their dot product and the logits are updated with the agreement measure.

We iterate through the proposed routing algorithm $i$ times, where we empirically set $i=3$ following \cite{sabour2017dynamic}. After the iterations, the output of the layer $l+1$, $\mathbf{\Phi}^{l+1}$ can be obtained by $\mathbf{\Hat{S}}$.

\newcommand*{\skipnumber}[2][1]{{\renewcommand*{\alglinenumber}[1]{}\State #2}\addtocounter{ALG@line}{-#1}}

\begin{algorithm}
\caption{Dynamic Routing using 3D convolution}
\label{alg:3d}
\begin{algorithmic}[1]
\Procedure{Routing}{}\\
\algorithmicrequire{ $\mathbf{\Phi}^{l} \in \mathbb{R}^{(w^{l},w^{l},c^{l},n^{l})}$, $r$ and $c^{l+1},n^{l+1}$}
\State $\mathbf{\Tilde{\Phi}}^{l} \gets \texttt{Reshape}(\Phi_{l})\ \in \mathbb{R}^{(w^{l},w^{l},c^{l}\times n^{l},1)}$
\State $\mathbf{V}  \gets \texttt{Conv3D}(\mathbf{\Tilde{\Phi}}^{l}) \in \mathbb{R}^{(w^{l+1},w^{l+1},c^{l},c^{l+1}\times n^{l+1})}$
\State $\mathbf{\Tilde{V}}  \gets \texttt{Reshape}(\mathbf{V})\ \in \mathbb{R}^{(w^{l+1},w^{l+1},n^{l+1},c^{l+1},c^{l})}$

\State $\mathbf{B} \gets \mathbf{0} \in \mathbb{R}^{(w^{l+1},w^{l+1},c^{l+1},c^{l})}$
\skipnumber{Let $p \in w^{l+1}, q \in w^{l+1}, r \in c^{l+1}$ and $s \in c^{l}$}
\For {$i$ iterations}
\State \text{for all }$p,q,r,$ $ \  k_{pqrs} \gets \texttt{softmax\_3D}(b_{pqrs})$
\State \text{for all }$s,$ $ \ S_{pqr} \gets \sum_{s} k_{pqrs} \cdot \Tilde{V}_{pqrs}$
\State \text{for all }$s,$ $ \ \hat{S}_{pqr} \gets \texttt{squash\_3D}(S_{pqr})$
\State \text{for all }$s,$ $ \ b_{pqrs} \gets b_{pqrs} + \hat{S}_{pqr} \cdot  \Tilde{V}_{pqrs}$
\EndFor
\State\Return $\mathbf{\Phi}^{l+1} = \mathbf{\hat{S}}$
\EndProcedure
\end{algorithmic}
\end{algorithm}

%% file: methods_deep_archi.tex
Even though the architecture proposed by \cite{sabour2017dynamic} performs well with MNIST, fashion MNIST~\cite{fmnist} and similar datasets, its performance on CIFAR10 and other datasets containing complex objects can be considered sub-par. This is due  to the fact that the MNIST images can be easily classified with low level features such as edges and blobs, while CIFAR10 images require high level understanding of features. Thus, in this paper we propose  a novel deep capsule architecture which contains 16 convolutional capsule layers and a fully-connected capsule layer. However, going deep with capsule networks poses several challenges, which we discuss and attempt to solve by proposing customized layers below.


In the first few layers of the network, as the feature map space is large, routing is computationally expensive at the start. Hence, we keep the number of routing iterations as one at the first few layers. We need to stack layers to build a deep capsule network. However, since all the operations are required to be in capsule form, stacking of convolutional layers will not be useful as it produces the outputs as scalar feature maps. Therefore, in order to address these requirements, we propose \texttt{ConvCaps} layer, which is similar to a convolutional layer, except that its outputs will be squashed 4D tensors. We use \texttt{ConvCaps} layer where $i=1$, and for any $i>1$ we use \texttt{ConvCaps3D} layer.

Let $\mathbf{\Phi}^{l} \in \mathbb{R}^{(w^{l},w^{l},c^{l},n^{l})}$ be the input to the \texttt{ConvCaps} layer and $\mathbf{\Phi}^{l+1} \in \mathbb{R}^{(w^{l+1},w^{l+1},c^{l+1},n^{l+1})}$ be the output from the layer $l$. $w^{l+1}$ is obtained from the convolutional strides and padding, refer (Eq. \ref{eq1}). First $\mathbf{\Phi}^{l}$ is reshaped into $(w^{l},w^{l},c^{l}\times n^{l})$ and convoluted with $(c^{l+1} \times n^{l+1})$ filters, producing $(c^{l+1} \times n^{l+1})$ feature maps of width and height $(w^{l+1},w^{l+1})$. This will then be reshaped into $(w^{l+1},w^{l+1},c^{l+1}, n^{l+1})$ shaped $\mathbf{\Phi}^{l+1}$ tensor and squash function is applied to the capsules. This helps us to convert the feature maps into the capsule domain. In \cite{sabour2017dynamic}, when $i=1$ the predictions are equally weighted sum of the votes. The convolution operation is an alternative way, except it gives a weighted sum of the input capsules to predict next layer votes. Further, when $i$ is set to a value greater than $1$,\ the $\texttt{ConvCaps3D}$ layer is used with 3D convolution based dynamic routing algorithm~\ref{alg:3d}.

In order to reshape \texttt{ConvCaps}, we introduce \texttt{FlatCaps}, which are used to remove the spatial relationship between adjacent capsules in \texttt{ConvCaps} layer $l$, while keeping the part-whole relationships between the capsules in \texttt{ConvCaps} layer $l$ and \texttt{FC\_caps} layer $l+1$. Thus, the \texttt{FlatCaps} takes a $(w^{l},w^{l},c^{l},n^{l})$ shaped tensor and reshape it into a $(a^{l}, n^{l})$ shaped matrix, where, $a^{l}=w^{l} \times w^{l} \times c^{l}$.

\texttt{FC\_caps} are similar to the fully connected layers in deep neural networks. Here, ${\Phi}^{l} \in \mathbb{R}^{(a^{l},n^{l})}$ is mapped into ${\Phi}^{l+1} \in \mathbb{R}^{(a^{l+1},n^{l+1})}$. Each capsule in $\Phi^{l}$ is transformed into a capsule in $\Phi^{l+1}$ by a transformation matrix $W_{i,j} \in \mathbb{R}^{n^{l} \times n^{l+1}}$. Here, the $W$s are learned during the training process via back propagation.

\begin{figure}[!h]
    \centering
    \includegraphics[scale=0.6]{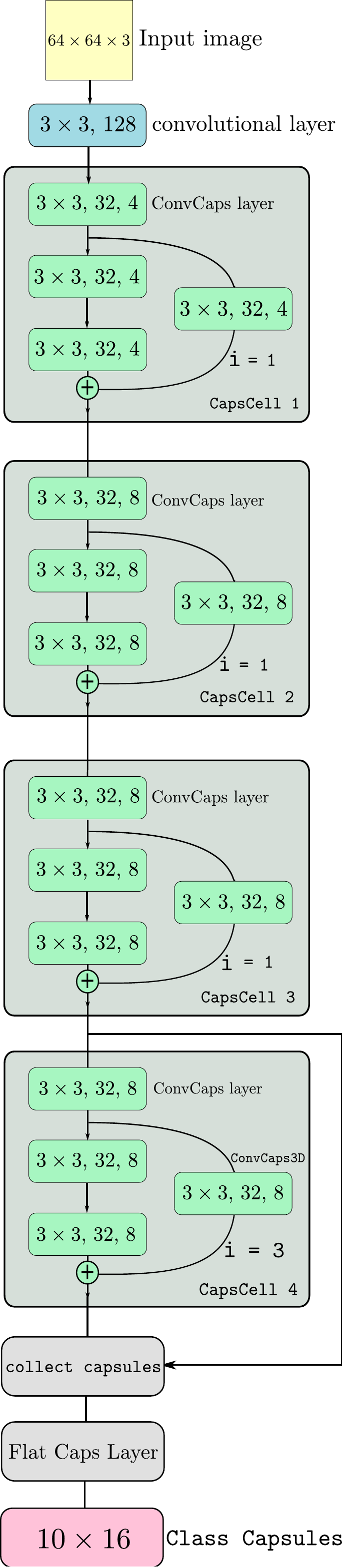}
    \caption{A four cell DeepCaps model, with first three cells using $i=1$ and in the last cell 3D convolution based dynamic routing is applied.}
    \label{fig:deepcaps}
    \vspace*{-0.5cm}
\end{figure}

With the use of these layers, we build our DeepCaps architecture as illustrated by Fig. \ref{fig:deepcaps}. The model contains four main blocks, skip connected CapsCells, 3D convolutional CapsCells, a fully-connected capsule layer and a decoder network. A skip connected capsule cell has three \texttt{ConvCaps} layers with the first layer output convolved and skip connected to the last layer output. The motivation behind skip connections is to reduce vanishing gradients in deep models.  In addition, this allows us to route low-level capsules to high-level capsules with skip
connections. We use element-wise layer addition to join the two capsule layers' outputs after the skip connections. Since the capsules are represented with vectors, a channel-wise concatenation was not used as it duplicates the same capsule, but element-wise addition reduces the bias and reduces the susceptibility to noise. Subsequently, we have a cell with \texttt{ConvCaps3D} layer, where the number of routing iterations is kept to 3. Then, the \texttt{ConvCaps} outputs are flattened and concatenated with the outputs of the capsules before 3D routing (in CapsCell 3) prior to the dynamic routing. Intuitively, this step aids to generalize the model for a broad range of diverse datasets. For an example, low level capsules from cell 1 or 2 would be sufficient for datasets consisting of images with poor information content such as MNIST, while we need to go deep enough until 3D ConvCaps capsules for datasets consisting of images with rich information content such as CIFAR10. Once all the capsules are collected and concatenated, they are routed to the class capsules via the \texttt{FC\_caps} layer. Here, the decision making happens and the input image is encoded into the final capsule vector. Finally, we use a decoder network to reconstruct the input image, as proposed in \cite{sabour2017dynamic}. However, the decoder proposed in \cite{sabour2017dynamic} merely consists of two fully connected layers, which cannot properly reconstruct the spatial relationships learned by the capsule network. Hence, we replace the decoder in \cite{sabour2017dynamic} with a deconvolutional decoder, which is better at reconstructing spatial relationships.

%% file: methods_deep_loss.tex
We use the margin loss \cite{sabour2017dynamic} as the loss function for DeepCaps. The marginal loss function enhances the class probability of the true class, while suppressing the class probabilities of the other classes. 

\begin{equation}
\begin{aligned}\label{eq:loss}
L_k &= T_k \max(0, m^+ - \|v_k\|)^2 \\
    &\ + \lambda (1-T_k)\max(0, \|v_k\| - m^-)^2 \\
\end{aligned}
\end{equation}

Here $T_k$ is 1 if the true class is $k$ and zero otherwise. We use $m^+ = 0.9$ and $m^- = 0.1$ as the lower bound for the correct class and the upper bound of the incorrect class as in Sabour \emph{et al.}~\cite{sabour2017dynamic}. $\lambda$ is used to control the effect of gradient back propagation at the initial phase of the training.

%% file: methods_decoder.tex
Our decoder network consists of deconvolutional layers \cite{zeiler2010decon} which reconstructs the input data by utilizing the instantiation parameters extracted from the DeepCaps model. In comparison with the fully-connected layer decoder~\cite{sabour2017dynamic}, this captures more spatial relationships while reconstructing the images. Further, we use binary cross entropy as the loss function for improved performance~\cite{8658735}.

The existing decoder, which is used as regularization for Capsule Networks, is class dependent. Let $\mathbf{P} \in \mathbb{R} ^{a \times b}$ contains the activity vector for all classes, where $a$ is the number of classes in final class capsule and $b$ is the capsule dimension. $\mathbf{P}$ is masked by the class with highest probability, results in $\mathbf{\hat{P}}$ as shown in below Eq. \ref{eq:mask}:

\begin{equation}
\begin{aligned}
\label{eq:mask}
\hat{p}_{i,j} = \begin{cases}
p_{i,j} & i = t\\
0 & i \neq t 
\end{cases}
\ \ \ \ 
\end{aligned}
\end{equation}

\begin{figure}[!htb]
\centering
\minipage{0.5\textwidth}
\centering
  \includegraphics[width=0.54\linewidth, angle=-90]{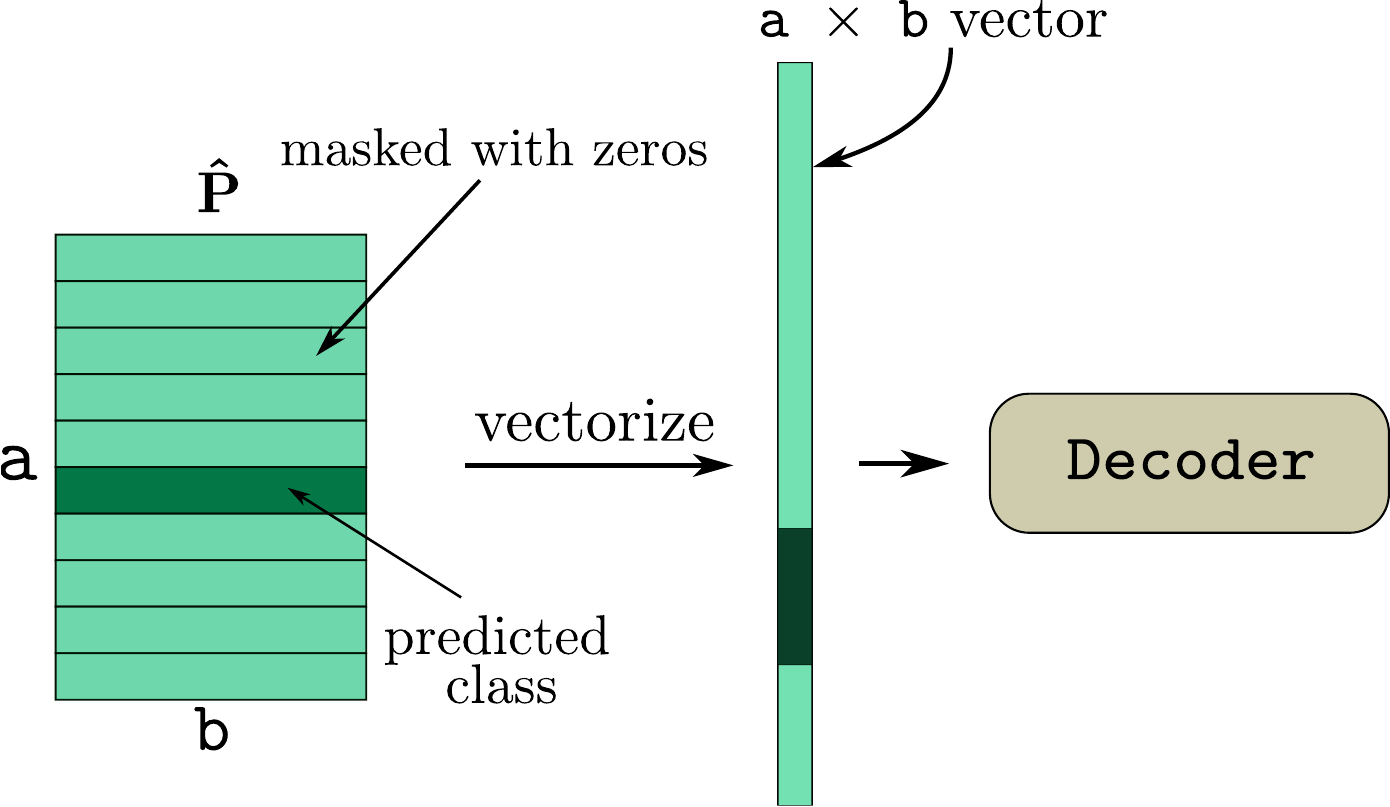}
  \vspace{0.2cm}
  \caption{Decoder network used in \cite{sabour2017dynamic}, which takes all the vectorized masked activity vectors.}
  \label{fig:dec}
\endminipage\\
\vspace{0.2cm}
\minipage{0.5\textwidth}
\centering
  \includegraphics[width=0.39\linewidth, angle=-90]{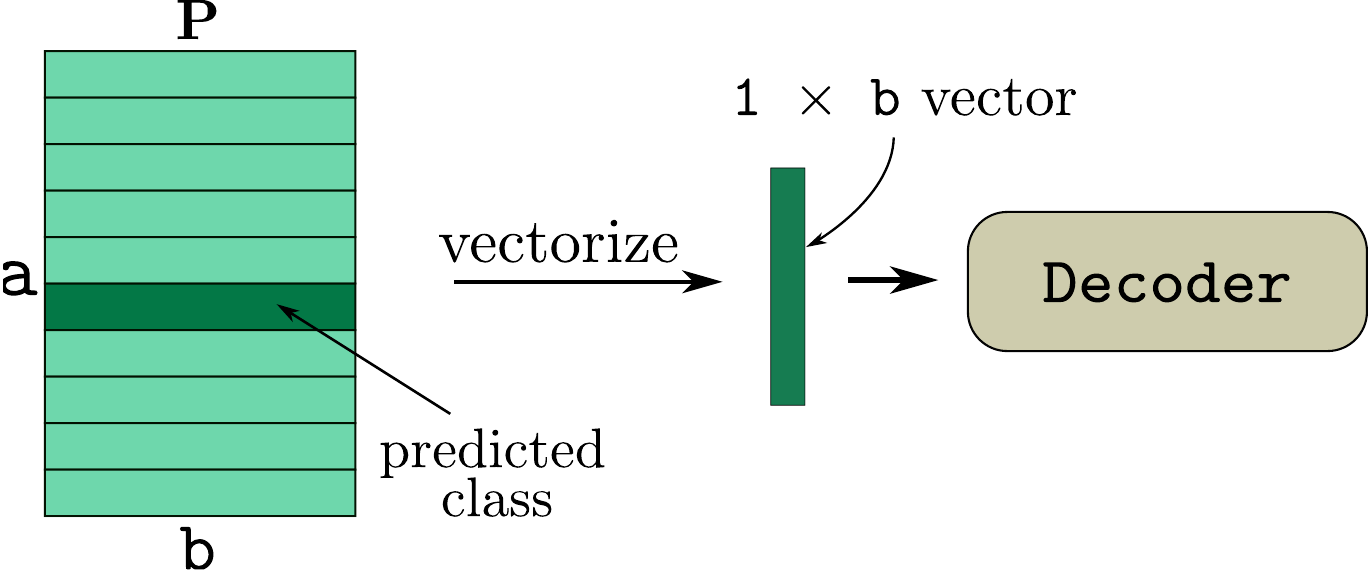}
  \vspace{0.2cm}
  \caption{Proposed decoder, which takes only the  activity vectors of the predicted class.}
  \label{fig:dec_new}
\endminipage\hfill
\end{figure}

Here $i \in [a], j\in[b]$  and $t = \mathrm{argmax}_i(\|\mathbf{P}_i\|^2_2)$ for the inference stage, and $t=true \ label$ in the training stage. Matrix $\mathbf{\hat{P}}$ is vectorized and fed in to the decoder network, as illustrated by Fig. \ref{fig:dec}. This vectorized $\hat{P} \in \mathbb{R}^{a \times b}$ contains non-zero values from $t \cdot b$ to $(t+1) \cdot b$ dimensions and zeros elsewhere. Therefore, the decoder network gets the class information from the dimension-specific distribution, which provides class information to the decoder indirectly, making the decoder class dependant. 

Hence, we propose a novel class-independent decoder network which acts as a better regularizer for the capsule networks, since it is forced to learn the activity vectors jointly within a constrained $\mathbb{R} ^{b}$ space. In our setting, only vector $P_{t} \in \mathbb{R}^{1 \times b}$ is fed into the decoder, where $t = \textit{true label}$ in the training stage, and $t = \mathrm{argmax}_i(\|\mathbf{P}_i\|^2_2)$.

Apart from regularization, a key advantage of having a decoder network is that it can be utilized for tasks such as data generation~\cite{ sabour2017dynamic}. 
However, a significant limitation of these decoders is the lack of controllability over which physical parameter is captured by which instantiation parameter. For example, if a certain instantiation parameter for a given class causes rotation for that particular class, there is no guarantee that the same instantiation parameter would cause rotation in any other classes. As a result, generating data with similar requirements, such as the same thickness or skewness, is a challenge. 

As a solution to these issues, we propose the following procedure. Instead of masking the non-predicted class instantiation parameters, we only send the $P_{t} \in \mathbb{R}^{1 \times b}$, as illustrated by Fig. \ref{fig:dec_new}. In contrast to the decoder learning procedure in \cite{sabour2017dynamic}, the learning of each instantiation parameter in the proposed method is drawn from the same joint distribution. Hence, the entity encapsulated by the any given instantiation parameter, which is learned by the decoder, will be the same irrespective of the image label.

Further, this procedure helps us to understand the types of variations in the MNIST dataset. For example, rotation and elongation being a dominant variation in the dataset while localized changes being less dominant among characters is reflected by the variance of the activity vector. In other words, the instatiation parameters causing rotations have higher variance whereas those causing localized changes have lower variance.

%% file: implementation.tex
We used Keras and Tensorflow libraries for the development of DeepCaps. For the training procedure, we used Adam optimizer \cite{kingma2015adam} with an initial learning rate of 0.001, which is reduced by half after each 20 epochs. During the initial phases of the training, $\lambda$ in Eq. \ref{eq:loss} is set to $0.2$ and increased to $0.5$ in the latter part of the training. The models were trained on GTX-1080 and V100 GPUs, and a weighted average ensembling was used for the 7-ensemble models reported in Table \ref{tbl:best}.

%% file: results.tex
\subsection{Classification Results}

We test our DeepCaps model with several benchmark datasets, CIFAR10 \cite{cifar10}, SVHN \cite{svhn} , Fashion-MNIST \cite{fmnist} and MNIST \cite{mnist}, and compare its performance with the existing capsule network architectures. For CIFAR10 and SVHN, we resize the $32 \times 32 \times 3$ images to $64 \times 64 \times 3$ and for other datasets, original image sizes are used throughout our experiments.

\begin{table}[!htb]
\caption{Classification accuracies of DeepCaps, CapsNet~\cite{sabour2017dynamic} and other variants of Capsule Networks, with the state-of-the-art results. We outperform all the capsule domain networks in CIFAR10, SVHN and Fashion-MNIST datasets, while achieving similar performace on the MNIST dataset.}
\begin{center}
\label{tbl:best}
\begin{adjustbox}{width=1\linewidth}
\begin{tabular}{|l|c|c|c|c|c|}
\hline
Model  & \footnotesize{CIFAR10} & \footnotesize{SVHN} & \footnotesize{F-MNIST} & \footnotesize{MNIST}  \\
\hline\hline
DenseNet~\cite{huang2017densely} &   96.40\% & 98.41\%   &  95.40\%   &  - \\
ResNet~\cite{DBLP:journals/corr/HeZRS15} & 93.57\%   &  - & -   &  99.59\% \\
DPN~\cite{DBLP:journals/corr/ChenLXJYF17} &  96.35\% &  - & 95.70\%   &  - \\
Wan \textit{et al.}~\cite{wan2013reg} &  - & -   &  -   & 99.79\%\\
Zhong \textit{et al.}~\cite{zhong2017random}& 96.92\% & -   & 96.35\% & -\\
\hline
Sabour \emph{et al.}~\cite{sabour2017dynamic}&  89.40\%     & 95.70\%  &   93.60\%      &  \textbf{99.75}\%\\
Nair \emph{et al.}~\cite{Nair2018PushingTL}&   67.53\%       & 91.06 \% &   89.80\% &  99.50\%\\
HitNet~\cite{deliege2018hitnet}&  73.30\%      & 94.50\%    &  92.30\% &  99.68\%\\
DeepCaps & 91.01\%     & 97.16\% &  94.46\% & 99.72\%\\
DeepCaps (7-ensembles) & \textbf{92.74}\%     & \textbf{97.56}\% &  \textbf{94.73}\% &  - \\
\hline
\end{tabular}
\end{adjustbox}
\end{center}
\vspace*{-0.5cm}
\end{table}

\begin{figure*}[h]
\minipage{0.5\textwidth}
\centering
  \includegraphics[width=\linewidth]{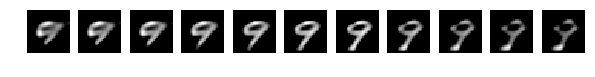}
\endminipage\hfill
\minipage{0.5\textwidth}
\centering
  \includegraphics[width=\linewidth]{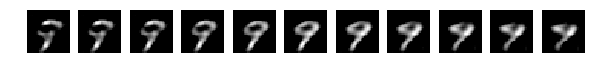}
\endminipage\hfill

\vspace{-0.4cm}
\minipage{0.5\textwidth}
\centering
  \includegraphics[width=\linewidth]{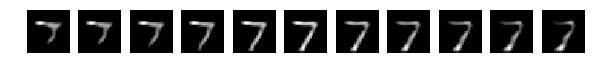}
\endminipage\hfill
\minipage{0.5\textwidth}
\centering
  \includegraphics[width=\linewidth]{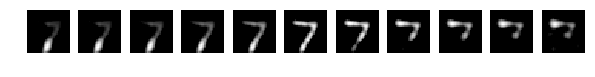}
\endminipage\hfill

\vspace{-0.4cm}
\minipage{0.5\textwidth}
\centering
  \includegraphics[width=\linewidth]{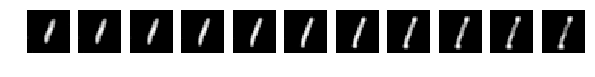}
\endminipage\hfill
\minipage{0.5\textwidth}
\centering
  \includegraphics[width=\linewidth]{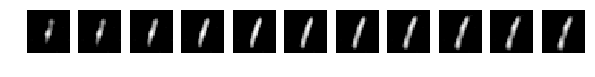}
\endminipage\hfill
\vspace{-0.4cm}

\minipage{0.5\textwidth}
\centering
  \includegraphics[width=\linewidth]{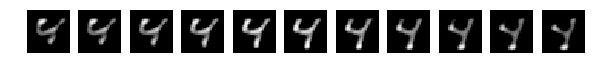}
\endminipage\hfill
\minipage{0.5\textwidth}
\centering
  \includegraphics[width=\linewidth]{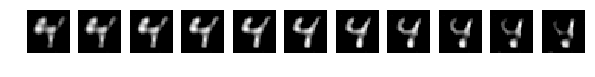}
\endminipage\hfill

\vspace{-0.4cm}

\minipage{0.5\textwidth}
\centering
  \includegraphics[width=\linewidth]{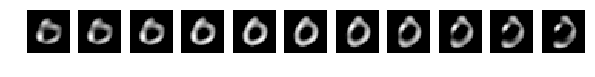}
\endminipage\hfill
\minipage{0.5\textwidth}
\centering
  \includegraphics[width=\linewidth]{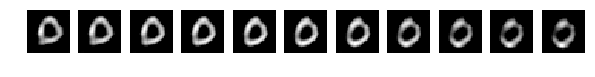}
\endminipage\hfill

\caption{Left half of images are generated by our decoder network, and the right half of the images are generated by decoder used in \cite{sabour2017dynamic}. When the $28^{th}$ dimension of the activity vector is changed between [-0.075,0.075], we can clearly observe that all the variations in the left half of the images are the same, like elongation in vertical direction. In the right half images, the variations are different for each class. For an example `7' is shrunken in the vertical dimension, `1' is elongated in the vertical direction and `9' is showing some rotation.}
\label{fig:recon_bold}
\end{figure*}

Even though our results are slightly below or on-par with the state-of-the-art results, our results comfortably surpass all the existing capsule network models in CIFAR10, SVHN and Fashion-MNIST datasets. If we take the capsule network implementations with best results, there is a 3.25\% improvement in CIFAR10 and 1.86\% improvement in SVHN compared to the capsule network model proposed in \cite{sabour2017dynamic}. For Fashion-MNIST dataset, we outperform the results of HitNet \cite{deliege2018hitnet} by 1.62\% and for MNIST, DeepCaps produced on-par state-of-the-art results. Table \ref{tbl:best} shows our results in comparison with the existing capsule network results and state-of-the-art results for the corresponding datasets. We highlight that we were able to achieve a near state-of-the-art performance across the datasets while surpassing the results of all the existing capsule network models.

We rescaled the images only for CIFAR10 and SVHN datasets as a data augmentation, since they have richer high-level features compared to MNIST and F-MNIST. Having $64 \times 64$ resolution images allows us to add more layers to go down deep in the network.

For the models trained on CIFAR10, DeepCaps has only 7.22 million parameters, while CapsNet \cite{sabour2017dynamic} has 22.48 million parameters. Still we achieved 91.01\% on CIFAR10 with a single model, where CapsNet has a 7 ensembles accuracy of 89.40\%. We tested both models' inference time on NVIDIA V100 GPU, CapsNet takes 2.86 ms for a $32 \times 32 \times 3$ image, while our model takes only 1.38 ms for a $64 \times 64 \times 3$ image.

\subsection{Class-Independent Decoder Image Reconstruction}

Our class-independent decoder acts as a better regularization term, yet it also helps to jointly learn the inter class reconstruction. Hence, all the instantiation parameters are distributed in the same space. For example, specific variations in the handwritten digit, such as boldness, rotation and skewness are captured at the same locations of the instantiation parameter vector for all the classes. In other words, for class `9' if the $7^{th}$ instantiation parameter is responsible for rotation, then it will be the same $7^{th}$ parameter causing rotation in any other classes as well. The outputs of the decoder used in \cite{sabour2017dynamic} is also subjected to changes in the perturbation of activity vectors, yet, a specific instantiation parameter may cause rotation in the reconstructed output for one class, and at the same time, it will not be the same instantiation parameter causing rotation in another class. This is due to the fact that the activity vectors are distributed in a dimensional-wise separable activity vector space. Using our class-independent decoder, we can generate data for any class with a certain requirement. For example, if we want to generate bold data from a text, once we find the instantiation parameter responsible for the boldness for any class, then we can perturb it to generate bold letters across all classes, which we can not do in \cite{sabour2017dynamic}, unless we know all the locations of instantiation parameters corresponding to boldness for all the classes. See Fig. \ref{fig:recon_bold}.

\begin{figure}
\minipage{0.48\textwidth}
    \centering
    \includegraphics[scale=0.55, angle = -90]{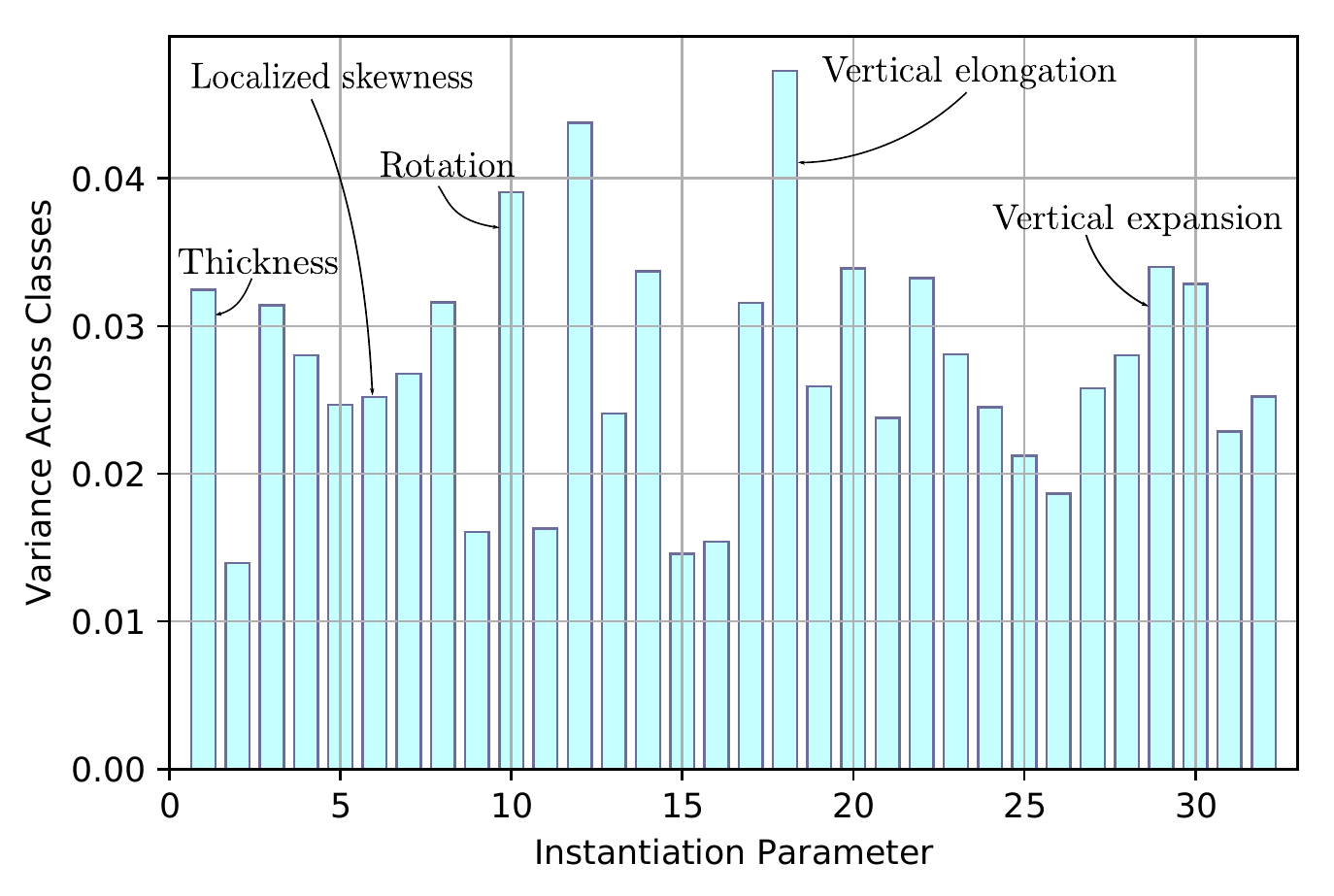}
    \caption{All the 32 instantiation parameters and its variance across the MNIST dataset. Although, instantiation parameter space is not orthogonal, high variance instantiation parameters show clear separable changes in the reconstructed images, while, low variance instantiation parameters show mixed changes.}
    \label{fig:graph}
\endminipage\\

\minipage{0.48\textwidth}
    \centering
    \includegraphics[scale=0.325, angle = -90]{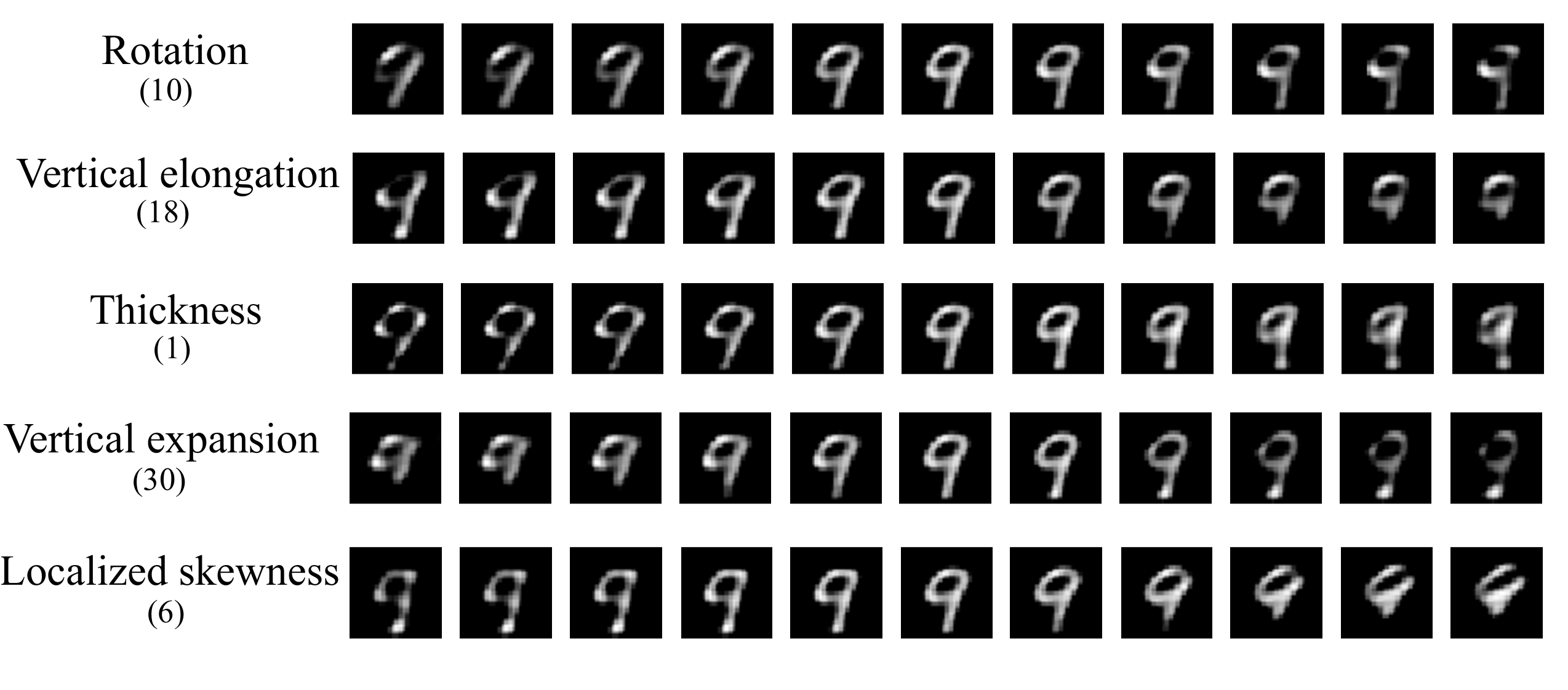}
    \caption{Perturbations on a single instantiation parameter of the above digit shows that, high variance instantiation parameters cause global changes and low variance instantiation parameters cause localized changes.}
    \label{fig:9}
\endminipage
\end{figure}

With this class-independent decoder, we can label each instantiation parameter causing specific changes in the reconstructed images. For the models that we trained, we observed that the $28^{th}$ parameter always causes the vertical elongation, and the $1^{st}$ parameter is responsible for thickness. Further, we observed that, when we rank these instantiation parameters by variance, the instantiation parameters with the higher variance causes global variations such as rotation, elongation and thickness, while parameters with lower variance are responsible for localized changes. See Fig. \ref{fig:graph}. The instantiation parameter space is not restricted to be orthogonal, hence, few instantiation parameters share a common attribute of an image. Yet, the instantiation parameters with higher variance demonstrates clearly separable variations as illustrated by Fig. \ref{fig:9}.


%% file: conclusion.tex
\section{Conclusion}

In this paper, we proposed a new deep architecture for Capsule Networks, termed DeepCaps, drawing intuition from the concepts of skip connections and 3D convolutions. Skip connections within a capsule cell allow good gradient flow in back propagation. At the bottom of the network, we use a higher number of routing iterations when the skip connections jump more than a layer. 3D convolutions are used to generate votes from the capsule tensors which are used for dynamic routing. This helps us to route a localized group of capsules to a certain higher-level capsule. As a result, we were able to go deeper with capsules using less computational complexity compared to Sabour \emph{et al.}~\cite{sabour2017dynamic}. Our model surpasses the state-of-the-art performance on CIFAR10, SVHN and Fashion-MNIST, while achieving the state-of-the art performance on MNIST datasets in the Capsule Network domain.

Further, we introduced a novel class-independent decoder network, which acts as a regularization for the DeepCaps. Since it learns from the activity vectors which are distributed in the same space, we observed that across all the classes, a specific instantiation parameter captures a specific change. This opens up new avenues in practical applications such as data generation.


Furthermore, we were able to get better performance on comparatively complex datasets such as CIFAR10, where the CapsNet in \cite{sabour2017dynamic} did not show significant performance. As future work, we would like to build even deeper and higher level understanding models and apply on ImageNet dataset. The class-independent decoder network also showed potential in data generation applications with specific requirements such as generate text data with same styles. Further, we hope to investigate on eliminating the correlation between the instantiation parameters.